%% file: main.tex
\documentclass{article}
\pdfoutput=1
\PassOptionsToPackage{numbers}{natbib}

\usepackage{multirow}
\usepackage[preprint]{neurips_2021}

\usepackage[utf8]{inputenc} 
\usepackage[T1]{fontenc}    
\usepackage{hyperref}       
\usepackage{url}            
\usepackage{booktabs}       
\usepackage{amsfonts}       
\usepackage{nicefrac}       
\usepackage{microtype}      
\usepackage{xcolor}         
\usepackage{amsmath}
\usepackage{amsthm}
\usepackage{algorithm}
\usepackage{algpseudocode}
\usepackage{caption}
\usepackage{graphicx}
\usepackage{epstopdf}
\usepackage{adjustbox}
\usepackage{graphicx}
\usepackage{subfig}
\usepackage{siunitx}
\usepackage{dsfont}

\theoremstyle{definition}
\newtheorem{definition}{Definition}[section]
\newtheorem{theorem}{Theorem}
\newtheorem{innercustomthm}{Theorem}
\newenvironment{customthm}[1]
  {\renewcommand\theinnercustomthm{#1}\innercustomthm}
  {\endinnercustomthm}

\newtheorem{corollary}{Corollary}
\newtheorem{innercustomcorollary}{Corollary}
\newenvironment{customcorollary}[1]
  {\renewcommand\theinnercustomcorollary{#1}\innercustomcorollary}
  {\endinnercustomcorollary}
\newtheorem{lemma}{Lemma}
\newtheorem{innercustomlemma}{Lemma}
\newenvironment{customlemma}[1]
  {\renewcommand\theinnercustomlemma{#1}\innercustomlemma}
  {\endinnercustomlemma}
\title{Efficient Inference via Universal LSH Kernel}

\author{%
  Zichang Liu\\
  Department of Computer Science\\
  Rice University\\
  Houston, TX 77025 \\
  \texttt{zichangliu@rice.edu} \\
  \And
  Benjamin Coleman\\
  Department of Electrical and Computer Engineering\\
  Rice University\\
  Houston, TX 77025 \\
  \texttt{ben.coleman@rice.edu} \\
  \And
  Anshumali Shrivastava\\
  Department of Computer Science\\
  Rice University\\
  Houston, TX 77025 \\
  \texttt{anshumali@rice.edu} \\
}

\begin{document}

\maketitle
\input{abstract}
\input{intro}

\input{relatedwork}

\input{method}

\input{experiment}
\input{conclusion}

\newpage
\bibliography{reference}
\bibliographystyle{unsrt}
\newpage
\newpage
\appendix
\input{appendix}

\end{document}

%% file: abstract.tex
\vspace{-6mm}
\begin{abstract}
Large machine learning models achieve unprecedented performance on various tasks and have evolved as the go-to technique. However, deploying these compute and memory hungry models on resource constraint environments poses new challenges. In this work, we propose mathematically provable \textsc{Representer Sketch}, a concise set of count arrays that can approximate the inference procedure with simple hashing computations and aggregations. \textsc{Representer Sketch} builds upon the popular \emph{Representer Theorem} from kernel literature, hence the name, providing a generic fundamental alternative to the problem of efficient inference that goes beyond the popular approach such as quantization, iterative pruning and knowledge distillation. A neural network function is transformed to its weighted kernel density representation, which can be very efficiently estimated with our sketching algorithm. Empirically, we show that \textsc{Representer Sketch} achieves up to $114\times$ reduction in storage requirement and $59\times$ reduction in computation complexity without any drop in accuracy.

  
\end{abstract}

%% file: intro.tex
\vspace{-4mm}
\section{Introduction}
\label{sec:intro}



Deep neural networks have led to stunning achievements across many application domains, particularly image processing systems~\cite{Krizhevsky_imagenetclassification,he2016deep} and natural language processing~\cite{vaswani2017attention,sak2014long}. These unprecedented performance have led to the rapid adoption of neural networks by practitioners from other fields such as healthcare~\cite{10.1038/s41591-018-0316-z,pmlr-v136-sarkar20a}, oil and gas~\cite{araya2018deep,zheng2019applications}, manufacturing~\cite{li2018deep,wangma} and agriculture~\cite{kamilaris2018deep,kussul2017deep}. However, the power of neural networks comes at a considerable cost; state-of-the-art networks require a tremendous number of computing operations, large storage and memory bandwidth, and high energy consumption. 
Such high cost complicates the deployment of networks in resource constraint settings such as mobile systems, sensors, actuators, and embedded systems.

Power consumption is a primary concern for resource-constrained devices, which often run on batteries. Memory access is a crucial contributor to the energy cost. Because deep neural networks are too large to store locally, they must be fetched from an external memory such as DRAM. On a commercial 45nm process, DRAM access costs 1.3-2.6 nJ, at least $65\times$ more than a cache fetch(20pJ)~\cite{Horowitz201411CE}. Second, a neural network forward pass consists of a series of matrix multiplications and non-linear functions, resulting in a massive number of expensive floating-point multiplies. On a commercial 45nm process, a 32-bit floating-point multiply requires $4\times$ more energy than a floating-point add~\cite{Horowitz201411CE}.

\textbf{Efficient Inference:} Popular approaches such as pruning~\cite{e9ee2143e19d49cf9cbe8861950b6b2a, han2015learning,orseau2020logarithmic,chao2020directional}, quantization~\cite{hubara2016quantized,37631,han2016deep,courbariaux2016binarized} and knowledge distillation~\cite{hinton2015distilling} have been introduced to address these challenges. Pruning eliminates non-important connections in the network; quantization reduces the memory by decreasing the number of bits needed for each weight; knowledge distillation replaces a neural network with a smaller one. However, existing methods still follow the form of neural network functions, which naturally require expensive multiplications involving a large number of parameters.

\textbf{Our Proposal:} We propose a compressed data structure (\textsc{Representer Sketch}) that reduces the inference process to hash computations and aggregations. We \textit{sketch} the learned networks into a succinct data structure that sharply approximates the inference function. This surprising possibility is a result of combining the \emph{Representer Theorem}, a well-known result in functional analysis, with recent advances in Locality Sensitive Hashing (LSH) and sketches for kernel density estimation. In the first stage of \textsc{Representer Sketch}, we transform a neural network function into its weighted kernel sum representation. Exact evaluation of this kernel representation is still prohibitively expensive during inference. Therefore, in the second stage, we use an efficient sketching algorithm to efficiently estimate the desired weighted kernel density. The sketch compresses all the data samples into a few arrays of counters. Due to the nature of the sketch, we may accurately estimate the kernel sum using only a few counter lookups and aggregations. 

We argue that \textsc{Representer Sketch} is efficient for the following two reasons. First, the storage requirement is reduced to a small number of weighted counters. This reduction allows moving the model from DRAM to more local storage such as cache. Second, neural network inference involves many expensive matrix multiplications, while \textsc{Representer Sketch} uses much cheaper operations such as addition and subtraction. The reduction in memory and computation complexity also leads to a more energy-friendly inference algorithm.

\textbf{Contribution:} In this paper, our main contribution is a provable and practical end-to-end inference algorithm that produces a tiny sketch, solving inference with simple computations. Specifically, 
\begin{itemize}
    \setlength\itemsep{0em}
    \item We propose an efficient sketching algorithm based on Locality Sensitive Hashing (LSH) for weighted kernel density estimation with rigorous estimation error bounds in Section~\ref{sec:ingredient2}. 
    \item We propose a method to transform a neural network function to its weighted kernel representation. To approximate this quantity with our sketching algorithm, we constrain the kernel functions to be LSH kernels according to the theoretical developments in Section~\ref{sec:bridgingthegap}.
    \item We conduct a comprehensive evaluation of \textsc{Representer Sketch} in Section~\ref{sec:experiment}. We show that the sketch reduces the memory requirement by up to $114\times$ and computation complexity by $59\times$ without losing accuracy. In a head-to-head  comparison with pruning and knowledge distillation, our sketch achieves better accuracy at every memory budget.
\end{itemize}

%% file: relatedwork.tex
\vspace{-4mm}
\section{Related Work}
\subsection{Efficient Neural Network Inference}
Over-parameterized neural networks are often beneficial for training and modeling, but they lead to a waste of both computation and memory for the inference process~\cite{NIPS2013_7fec306d}. Various methods have been proposed to solve the costly inference problem, with three main directions. The first line of work focuses on reducing the number of connections by pruning neural networks~\cite{e9ee2143e19d49cf9cbe8861950b6b2a,han2015learning,orseau2020logarithmic,chao2020directional}. A similar idea is to replace the set of nodes in each layer with a \textit{coreset} (i.e. a ``core set'' of nodes that are sufficient for inference)~\cite{dubey2018coresetbased}. Pruning based methods rely on the observation that a small sub-network can achieve similar performance to the original over-parameterized neural network. The second line of work, quantization, attempts to represent the model parameters and activations using fewer bits~\cite{hubara2016quantized,37631,han2016deep,courbariaux2016binarized}. With the appropriate hardware-level support, quantization can greatly reduce the memory requirements and energy consumption, making it a popular choice for embedded deployments. The third and final direction is knowledge distillation, originally proposed by \cite{hinton2015distilling} in the context of neural network ensembles. Knowledge distillation algorithms~\cite{hegde2019variational,mirzadeh2019improved,sanh2020distilbert} train a small ``student model'' to mimic the behavior of a larger ``teacher model.'' Our method can be viewed as a knowledge distillation based method in terms of the learning procedure. However, instead of distilling the knowledge into a smaller neural network, we reduce inference to kernel density estimation via the Representer Theorem.




\subsection{Locality Sensitive Hashing}
\label{sec:rwlsh}
Locality-sensitive hash (LSH) functions are a key ingredient in our algorithm. An LSH function is a function that maps similar points to the same hash value with a high probability. The formal definition of a LSH family is as follows~\cite{10.1145/276698.276876}


\begin{definition} [\bf $(S_0,cS_0,p_1,p_2)$-sensitive hash family]
    \label{def:lsh}
A family $ \mathcal{H}$ is called $(S_0,cS_0,p_1,p_2)$-sensitive with respect to a similarity function $sim(\cdot,\cdot)$ if for any two points $x,y \in \mathbb{R}^{D}$ and $h$ chosen uniformly from $\mathcal{H}$ satisfies:
    \begin{itemize}
        \item If $sim(x,y) \ge S_0$ then ${Pr}[h(x) = h(y)] \ge p_1$
        \item If $sim(x,y)\le cS_0$ then ${Pr}[h(x) = h(y)] \le p_2$
    \end{itemize}
\end{definition}
The probability ${Pr}[h(x) = h(y)]$ is known as the collision probability of $x$ and $y$. By collision, we mean that the hash values for two points are equal. For the purpose of our arguments, we use a stronger notion of LSH, in which the collision probability is a monotonic function of the similarity between $x$ and $y$. Ths is 
\begin{equation}
    {Pr}[h(x) = h(y)] \propto f(sim(x,y))
\end{equation}

Under these conditions, the collision probability forms a positive semi-definite radial kernel~\cite{coleman2019sublinear}. Examples include MinHash for Jaccard distance, sign random projections for angular distance, and p-stable LSH scheme for the Euclidean distance.

\subsection{Repeated Array-of-Counts Estimator(RACE)}
\label{sec:rwrace}
Recent works have shown that LSH can be used for efficient statistical estimation~\cite{spring2017new}~\cite{charikar2018hashingbasedestimators}~\cite{10.1145/3178876.3186056}~\cite{coleman2020sublinear}. In particular, RACE~\cite{coleman2019sublinear} is a hash-based sketch that solves the kernel density estimation (KDE) problem on streaming data.
The RACE sketch is a 2D array of counters with $L$ rows and $R$ columns. 
For each point $x$ in the dataset $D$ and each row in the sketch, RACE uses an LSH function $h(x)$ to calculate an index and increment the corresponding counter in the array. 
The process is repeated for each of the $L$ rows. 
Every counter represents the number of elements in the dataset whose hash value is that particular index, and the result is a sketch that approximates KDE for a query. \cite{coleman2019sublinear} proved that RACE could estimate the KDE with tight error bounds given a sufficiently large number of repetitions, provided that the kernel is the collision probability of an LSH function. In this paper, we extend the RACE framework to handle \textit{weighted} kernel density estimation.

%% file: method.tex
\vspace{-2mm}
\section{Representer Sketch: A Novel Foundation of Efficient Inference for Any Machine Learning Model}

We begin by describing some known results. Our algorithm is motivated by the well-known \emph{Representer Theorem}, which states that model predictions may be expressed as a weighted sum of positive semi-definite kernels.
Recent methods from the sketching and hashing literature provide very efficient ways to estimate weighted kernel sums, but these methods only apply to a special subset of kernels - not the general, data-dependent kernels required by the Representer Theorem.
In this work, we seek to understand whether the model prediction function can be expressed in terms of these special kernels, so that the efficient methods can be used for inference.

\subsection{Ingredient 1: Representer Theorem}
\label{sec:ingredient1}
The \emph{Representer Theorem}~\cite{10.1007/3-540-44581-1_27} is a well-studied result that serves as the basis for many machine learning algorithms. Any function, minimized by an arbitrary error function with a regularizer, can be expressed as a kernel sum.
\begin{equation}
    \label{eq:representer_thm}
    \centering f_N(x) = \sum_{i=1}^{N}\alpha_i \mathcal{K}(x, x_i)
\end{equation}
$\mathcal{K}$ is a task-dependent kernel function, $\alpha_i \in \mathbb{R}$ is a weight associated with each training point $x_i \in \mathbb{D}$, $\mathbb{D}$ is the training dataset and $x_i \in \mathbb{R}^d$. The prediction function $f_N$ may come from any linear or non-linear machine learning model. In this paper, we consider $f_N$ as belonging to a neural network. Equation~\ref{eq:representer_thm} clearly implies that any learned function, including neural networks, can be expressed as a weighted kernel density over the training data.

\begin{figure}[t]
  \vspace{-4mm}
  \label{fig:diagram}
  \centering
    \includegraphics[width=\textwidth]{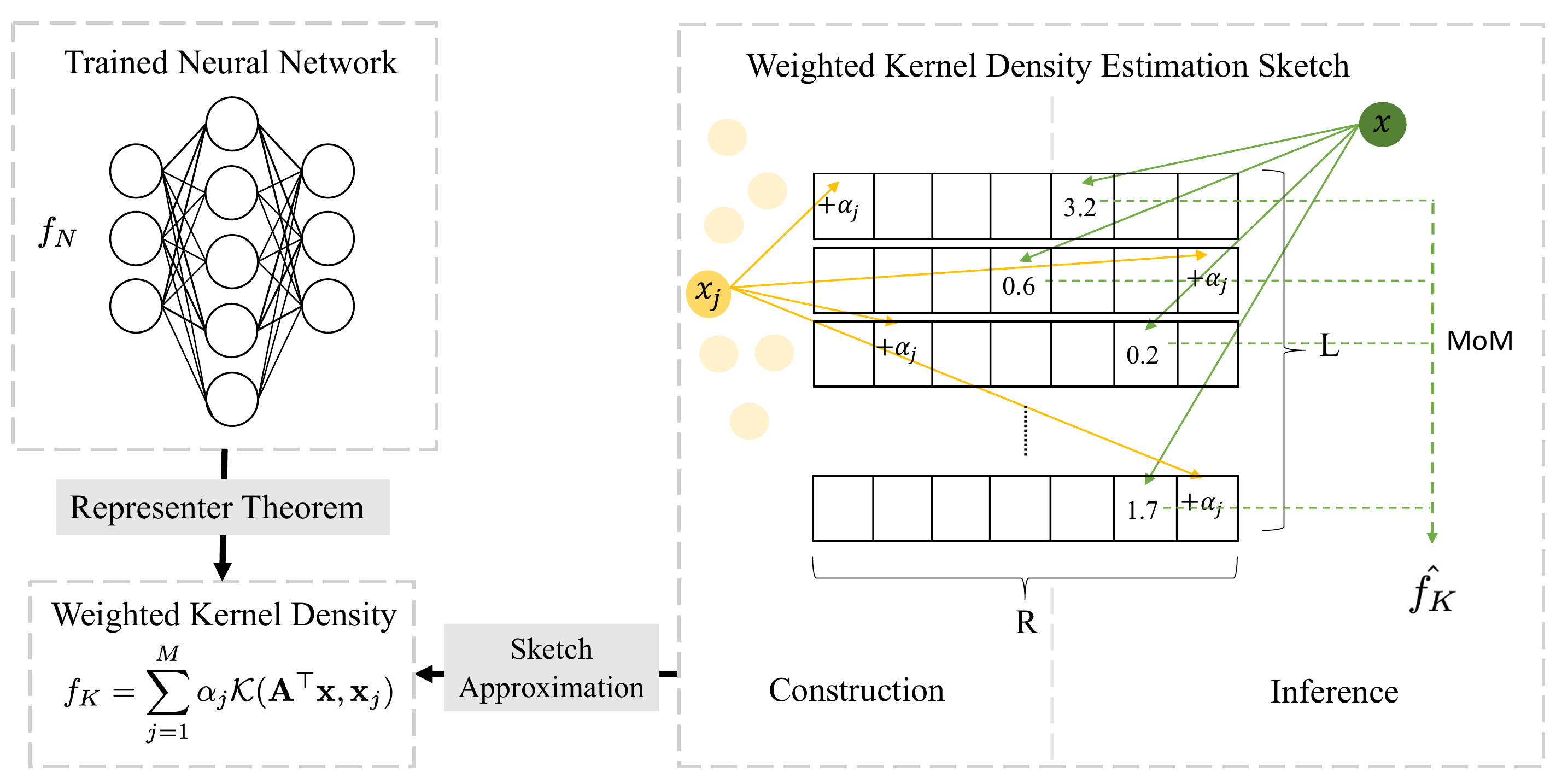}
  \caption{\textsc{Representer sketch} workflow: A trained neural network $f_N$ can be represented as a weighted kernel density function $f_K$.We obtain such representation using the output of $f_N$ as labels and mean square error as risk function. Then, we construct a sketch to approximate $f_K$ efficiently. The sketch is a 2d array of size $L \times R$. For every vector $x_j$, we calculate its index in every row and add its weight $\alpha_j$ to the valued store at the index. During inference, we calculate the index for testing data and retrieve $L$ values. The Median of Means (MoM) is returned as the inference result.}
  \vspace{-4mm}
\end{figure}

\subsection{Ingredient 2: Sketches for Weighted LSH Kernel Density Estimation}
\label{sec:ingredient2}





To efficiently estimate weighted kernel density sums, we propose a modification to the RACE sketch~\cite{coleman2019sublinear}. Given a dataset $\mathbb{U}$ of $M$ points $\{x_1, x_2, \dots, x_M\}$ and a kernel $\mathcal{K}$, the weighted kernel density estimate (KDE) of a query $q$ is defined as $\sum_{i=1}^{M}\alpha_i \mathcal{K}(q, x_i)$. Exact computation of the weighted KDE requires $O(M)$ kernel function evaluations. This is prohibitively expensive, so we propose a sketching algorithm that works when $\mathcal{K}$ is an LSH kernel (See Section~\ref{sec:rwlsh} for the definition). It should be noted that~\cite{coleman2020sublinear} consider a similar sketch, but ultimately only use weights $\alpha_i \in \{\mkern1.8mu{+}\mkern1.8mu 1,\mkern1.8mu{-}\mkern1.8mu 1\}$ and do not prove tight error bounds in terms of the weighted KDE. 

\textbf{Weighted KDE Sketch:} Our sketch $S$ is a 2D array with $L$ rows and $R$ columns. To add a point $x_i \in \mathbb{U}$ to the sketch, we calculate the index of $x_i$ in each row using an LSH function $h_l$ for $l = {1,\dots,L}$. Then, we increment the array at location $S[l, h_l(x_i)]$ with the weight value $\alpha_i$ associated with $x_i$. To query the
sketch, we hash the query $q$ with the same LSH functions to get a set of locations $\{h_1(q),... h_L(q)\}$. We output the mean of the array values at locations $S[l, h_l(q)]$. 

\textbf{Intuition:} The weighted kernel density measures the number of similar data points from the dataset, weighted by the relative importance of each point.
Because LSH functions cause similar points to collide, the LSH function used in our sketch will cause the query to map to the same array location as many similar points.
This array location contains the sum of weights for these similar points, which is a good approximation of the weighted KDE.

This sketch is an extension of the RACE sketch introduced in Section~\ref{sec:rwrace}. The key difference is that we use a floating point array to accommodate non-integral weights and that we increment each array value by $\alpha_i$ rather than 1. For a detailed description of the RACE sketching framework, see~\cite{coleman2019sublinear}.

\input{race_theory}



\subsection{Bridging the gap: From General Kernel to LSH Kernel Density}
\label{sec:bridgingthegap}
\input{rep_theory}

\subsection{The Whole Recipe}
\label{sec:wholerecipe}
\begin{algorithm}[t]
\begin{algorithmic}
    \State \textbf{Input:} \text{Dataset} $\mathbb{U},$ \text{Weights} $\{ \alpha_{i} \}_{i \in |\mathbb{U}|},$ \text{RACE rows} $L,$ \text{columns} $R$
    \State \textbf{Output:} sketch $S \in \mathbb{R}^{L \times R}$
    \State \textbf{Initialize:} $L$ independent LSH functions ${h_1,\dots,h_L}$
    \State $S \leftarrow 0^{L \times R}$
    \For{$x_i \in \mathbb{U}$}
        \For{ $l = 1 \rightarrow L$}
            \State $S[l, h_l(x_i)] += \alpha_i$
        \EndFor
    \EndFor
\end{algorithmic}
\caption{Representer Sketch Construction}
\label{alg:sketchbuild}
\end{algorithm}
\begin{algorithm}[t]
\begin{algorithmic}
    \State \textbf{Input:} Test data $q$, Sketch $S \in \mathbb{R}^{L \times R} $, number of means $g$
    \State \textbf{Output:} Approximate weighted kernel density output of $\hat{f_K}(q) = \sum_{j=1}^{M}\alpha_j\mathcal{K}(q, \mathbf{x}_j) $ 
    \State \textbf{Initialize:} $L$ independent LSH functions ${h_1,\dots,h_L}$ using the same random seed as in sketch construction
    \State Vector of means $M \leftarrow 0^{g}$
    \For{$i = 1 \rightarrow g$}
        \For{ $j = 1 \rightarrow L/g$}
            \State $l = i * L / g + m$
            \State $M[i] = M[i] + \frac{1}{L/g}S[l, h_l(q)]$
        \EndFor
    \EndFor
    \State $\hat{f_K}(q) \leftarrow$ median($M$)
\end{algorithmic}
\caption{Inference via Sketch}
\label{alg:inf}
\end{algorithm}
In this section, we present the whole workflow of \textsc{Representer Sketch}.
In Section \ref{sec:bridgingthegap}, we showed that given a neural network function $f_N$, there is an LSH weighted kernel density representation $f_K$ in the following form: 
\begin{equation}
    \label{eq:kernel}
    \centering f_K(\mathbf{q}) = \sum_{j=1}^{M}\alpha_j\mathcal{K}(\mathbf{q}, \mathbf{x}_j)
\end{equation}
$\mathcal{K}$ is an LSH kernel function and $\mathbb{U}$ is a set of vectors $x_j \in \mathbb{R}^d$ (or $\mathbb{R}^{d'}$, using Corollary~\ref{cor:transform}). Each $x_j$ is associated with a weight $\alpha_j$. Thus, if we wish to learn this representation, the trainable parameters are the weights $\alpha$, and the vectors $x_j$. The hyperparameters are the number of points $M$ and the LSH kernel $\mathcal{K}$. We train $f_K$ to approximate the model output function $f_N$ using gradient descent. The output of $f_{N}$ is used as the target with mean square error as the risk function.



\textbf{Building the Representer Sketch:} Once we have learned the weights and points for $f_K$, we build a sketch to approximate $f_K$ based on Section~\ref{sec:ingredient2}. We construct the sketch using Algorithm~\ref{alg:sketchbuild}. To recap from Section~\ref{sec:ingredient2}, we add $\alpha_j$ to the array locations identified by the LSH values of each $x_j \in \mathbb{U}$. 
In our experiments, it should be noted that each LSH function $h_l$ is constructed by concatenating $K$ independent LSH functions. The output of concatenated hash function $h_l$  is the set $[h_{l,1}(x),... h_{l,K}(x)]$, mapped to $\mathbb{Z}$ using a suitable transformation.


\textbf{Inference with the Representer Sketch:} Given a query $q$, we wish to evaluate $f_N$ during inference. This is done by using the sketch to estimate $f_K$. We present the process in Algorithm\ref{alg:inf}. We calculate $q$'s index for every row in the sketch, retrieve the values stored at every index, and return their mean as the estimates.


\textbf{Memory Requirement:} For inference, we need to store the sketch and a random seed to generate hashing functions. The sketch consists of $L * R$ floating numbers. Combining Theorem~\ref{thm:mem_error} and Theorem~\ref{thm:lshrepresenter}, we can obtain a relationship concerning the sketch memory and the estimation error between \textsc{Representer Sketch} and neural network functions. In general, more rows lead to smaller errors. In practice, $L$ can usually be smaller than 2000 and $R$ less than 20 for good results. The sketch would take less than 1 Mb and easy to store in the local cache.

\textbf{Computation Requirement:} From a computation perspective, the inference process mainly consists of two stages: calculating hash codes and sketch lookup. In practice, hash computations are known for being inexpensive. There are in total $K * L$ hashing functions. We further exploit the well-known method proposed by~\cite{ACHLIOPTAS2003671}. The parameters of each hash function are randomized from ${-1, 0, 1}$ and $\frac{2}{3}$ of them are zeros. This enables us to reduce the hash computation to only addition and subtraction. Calculating the hash codes becomes trivial to parallel and easy to implement in hardware for ultra-fast speed. And since the sketch is small enough to store in the local cache, array lookup is fast as well.

\textbf{Energy Requirement:}  Energy consumption is usually made of two parts: the number of bits accessed at different levels of memory and the number of floating points operation. Energy is mainly dominated by memory access, as we discussed in Section~\ref{sec:intro}. Memory access from external memory is much more energy expensive than from local memory. Thus, memory reduction would enable us to store the sketch into cache and significantly reduce energy consumption from memory access. Besides, \textsc{Representer Sketch} mainly rely upon addition and subtraction, and these operations are much cheaper in energy consumption comparing to multiplication.

%% file: race_theory.tex
\subsubsection{Theoretical Analysis}
In this section, we will show that our sketching algorithm yields a sharp unbiased estimator for weighted-KDE$(x)$. We defer proofs to the appendix.
\label{sec:theory}


\begin{theorem}[Sketch Estimator]
\label{thm:expandvar}
Given a dataset $\mathcal{D}$ of weighted samples $\{(\alpha_{x_i},x_i)\}$, let $h(x)$ be an LSH function drawn from an LSH family with collision probability $\mathcal{K}(\cdot,\cdot)$. Let S be a row of the sketch constructed using $h(x)$. For any query $q$,
\begin{equation*}
    \mathbb{E}(S[h(q)]) = \sum_{i}^{|\mathcal{D}|} \alpha_{x_i} \mathcal{K}(x_i,q), \qquad
    \mathrm{var}\left(S[h(q)]\right) \leq \left( \sum_{i}^{|\mathcal{D}|} \alpha_{x_i} \sqrt{\mathcal{K}(x_i,q)} \right)^2
\end{equation*}
\end{theorem}

The key insight is that each row of the sketch is an unbiased estimator for weighted sums of LSH kernels. We can obtain a sharp concentration around the weighted kernel sum by averaging these $L$ estimators. 
Although the simple average is often used in practice to query the sketch, we analyze the median-of-means, a more robust way to estimate the mean of $L$ i.i.d. random variables.
The median-of-means algorithm allows us to prove an exponential concentration of the estimate around the weighted KDE. Equivalent guarantees cannot be shown for the average without bounded or sub-Gaussian assumptions.
Furthermore, empirical results show that the average and median-of-means estimates have comparable performance. 
We use a result from~\cite{alon1999space}, where we suppose for convenience that $g$ evenly divides $N$.
\begin{lemma}
\label{lem:med}
Let $Z_1,...Z_R$ be $L$ i.i.d. random variables with mean $\mathbb{E}[Z] = \mu$ and variance $\leq \sigma^2$. Divide the $L$ variables into $g$ groups so that each group contains $m = L / g$ elements, and take the empirical average within each group. The median-of-means estimate $\hat{\mu}$ is the median of the $g$ group means. If $g = 8 \log(1/ \delta)$ and $m = L / g$, then the following statement holds with probability $1 - \delta$.
$$ |\hat{\mu} - \mu|\leq 6\frac{\sigma}{\sqrt{L}} \sqrt{\log{1/\delta}} $$


\end{lemma}

We combine the variance bound from Theorem~\ref{thm:expandvar} with the median-of-means guarantee to obtain a relationship between the sketch parameters, weighted kernel values, and estimation error. 

\begin{theorem}[Weighted-KDE Estimation Error]
\label{thm:mem_error}
Let $Z(q)$ be the median-of-means estimate constructed using the $L$ unbiased estimators of the sketch with $L$ rows. Then with probability $1 - \delta$,

\begin{equation*}
    |Z(q) - f_K(q)| \leq 6\frac{ \tilde{f}_K(q)}{\sqrt{L}}\sqrt{\log{1/\delta}}
\end{equation*}
where $f_K(q)$ and $\tilde{f}_K(q)$ are the weighted KDE with kernels $\mathcal{K}(x,q)$ and $\sqrt{\mathcal{K}(x,q)}$, respectively.
\end{theorem}

%% file: rep_theory.tex
\label{sec:33}

In this section, we show that a weighted LSH kernel sum is sufficient to approximate any continuous and bounded function, including that of a neural network. In contrast to the \emph{Representer Theorem}, which requires an unconstrained choice of kernel, we consider the more difficult setting where the kernel is a preselected LSH kernel. Our results show an interesting symmetry about the problem: if we constrain the KDE points to be the training datset, we must have a flexible kernel. If we constrain the kernel, we must have flexible point locations.


The crucial property is that LSH kernels are  \textit{universal}. Informally, a kernel is universal if a weighted kernel sum over a carefully-chosen point set can approximate any function with arbitrary accuracy~\cite{micchelli2006universal}. Some restrictions apply; the size of the point set is allowed to be arbitrarily large, the accuracy is computed over any compact subset of $\mathbb{R}^{d}$ rather than the entire space, and -- most critically -- the points in the point set may adapt to the function and are not necessarily taken from the training data.

\begin{definition} \textit{Universal Kernel}~\cite{micchelli2006universal}: Let $S(\mathcal{X})$ denote the space of bounded and continuous functions on a compact domain $\mathcal{X}$. Then a kernel $\mathcal{K}(x,y)$ is universal if it is continuous and induces a reproducing kernel Hilbert space that is dense in $S(\mathcal{X})$.
\end{definition}
Applied to our problem, a universal kernel is one which can approximate any well-behaved function over compact subsets of $\mathbb{R}^d$ using a linear combination of kernels. The theory of universal kernels allows us to construct a ``representer-style'' theorem for weighted LSH kernel sums. We begin by showing that the $p$-stable LSH functions of
are universal kernels.


\begin{lemma} 
\label{lemma2}
The L2 LSH kernel from~\cite{coleman2019sublinear} is shift-invariant and universal.
\end{lemma}

\begin{theorem} Given a continuous and bounded function $g(q)$ and $\epsilon > 0$, there exists a set of coefficients $\{\alpha_n\}$, set of points $\{x_n\}$ and an integer $N$ such that 
$$f_N(q) = \sum_{n=1}^N \alpha_N \mathcal{K}(x_n,q) \qquad \|f_N(q) - g(q)\|_\mathcal{X} \leq \epsilon$$
where $\mathcal{K}(x_n,q)$ is the L2 LSH kernel and $\mathcal{X}$ is any compact subset of $\mathbb{R}^{d}$.
\label{thm:lshrepresenter}
\end{theorem}

A consequence of Theorem~\ref{thm:lshrepresenter} is that we may perform any injective feature transformation of the query. In this case, we can construct the weighted kernel sum in the transformed space.

\begin{corollary}
\label{cor:transform}
Given an injective transform $T(q): \mathbb{R}^{d} \mapsto \mathbb{R}^{d'}$ and a continuous and bounded function $g(q): \mathbb{R}^{d} \mapsto \mathbb{R}$, there exists a set of points $\{x_n\}$ in the \textit{transformed space} $\mathbb{R}^{d'}$, coefficients $\{\alpha_n\}$, and an integer $N$ such that 
$$f_N(q) = \sum_{n=1}^N \alpha_n \mathcal{K}(x_n,T(q)) \qquad \|f_N(q) - g(q)\|_S \leq \epsilon$$
\end{corollary}

%% file: experiment.tex
\vspace{-3mm}
\section{Experiments}
\label{sec:experiment}
Our goal in this section is to evaluate \textsc{Representer Sketch} as a general efficient inference algorithm. We provide two sets of experiments. First, we validate our proposal of  \textsc{Representer Sketch} and contrast its accuracy and performance compared to the the trained neural network model. Specifically, we quantify the memory, computations, and accuracy tradeoffs with \textsc{Representer Sketch}.  In the second experiments, we compare \textsc{Representer Sketch} with popular iterative pruning methods~\cite{han2015learning} and knowledge distillation~\cite{hinton2015distilling} on the full memory accuracy trade-off curve. In particular, we are interested in answering whether \textsc{Representer Sketch} maintains high accuracy at various memory budget compared to baselines.

\subsection{Datasets}
We want to evaluate Representer Sketch as a general method that can be applied to various tasks from various domains.
From UCI datasets\cite{Dua:2019} on libsvm website\cite{CC01a}, we choose {\bf all} binary classification dataset with more than 10,000 data samples and {\bf all} regression dataset with more than 1000 data samples. Libsvm provides a standard data format. No feature engineering is performed on any dataset. Table~\ref{table:dataset} contains information about our datasets. These datasets covers a wide variety of tasks and domains. 

\subsection{Baselines}
\label{sec:baseline}

We consider iterative pruning~\cite{han2015learning} and knowledge distillation~\cite{hinton2015distilling} as our baseline for comparing the trade-off between memory and accuracy. For pruning, we choose global iterative pruning, zeroing out the lowest L1-norm connections across the whole model. One-Time Pruning refers to pruning and fine-tuning; Multi-Time Pruning refers to pruning and fine-tuning multiple times. We use the implementation from PyTorch. For knowledge distillation, we follow \cite{hinton2015distilling} and refer to this implementation~\cite{kd-zoo}. We do not include quantization as it is complementary and has limitation on compression rate. Quantization achieves at most $32\times$ reduction(reduce 64bit float into 2bit) and $2-4\times$ reduction in most case(PyTorch only provides implementation supporting 8-bit quantization).

\begin{table*}[t]
\caption{We summarize the accuracy, memory, and computation requirement in this table. Accuracy measures the classification accuracy for the top four datasets and mean absolute error for the bottom two datasets. NN refers to a converged neural network, Kernel refers to its weighted kernel density function, and RS refers to Representer Sketch. Kernel and RS achieve at least competing accuracy on all datasets. RS reduces memory footprint up to 114x, bringing the model size down to the scale of KB. RS reduces the number of floating operations by up to 59x.}
\centering
\begin{adjustbox}{max width=\textwidth}
\begin{tabular}{c||ccc|ccc|ccc}
    \hline
    \multirow{2}{4em}{ Datasets } & \multicolumn{3}{c|}{Accuracy} & \multicolumn{3}{c|}{Memory(MB)} & \multicolumn{3}{c}{FLOPs}\\
    \cline{2-10}
         & NN & Kernel & RS & NN & RS & Reduction & NN & RS & Reduction\\
    \hline
    Adult & 0.820 & 0.829 & 0.829  & 1.82  &  0.016 & 114x  & 0.227M   & 3.8K & 59x\\
    \hline
    phishing & 0.954 & 0.954 & 0.954  & 1.60  &  0.031 & 51x  & 0.199M  & 9.8K & 20x\\
    \hline
    skin & 0.999 & 0.997 & 0.997  & 0.338  &  0.019 & 17.8x  & 41.78K  & 3.97K & 11x\\
    \hline
    SUSY & 0.803 & 0.802 & 0.790  & 5.73  &  0.41 & 69x & 0.715M  & 0.177M & 4x\\
    \hline
    abalone & 1.51 & 1.52 & 1.51  & 0.28  & 0.006  & 46x & 34.94M  & 2.38K  & 14x \\
    \hline
    YearMSD & 12.06 & 12.05 & 11.24  & 6.25  & 0.12  &  50x & 0.78M   & 87.5K  & 10x \\
    \hline
	\end{tabular}
	\end{adjustbox}
	\label{table:mainresult}
	\vspace{-4mm}
\end{table*}

\begin{table*}[t]
\caption{We summarize the dataset information and parameter settings in this table. RS parameters correspond to the results in Table\ref{table:mainresult}. We use MLP as the neural network architecture.NN parameters show the hidden size of each hidden layer. When possible, we choose the number of layers and hidden layer size according to the original paper proposing the dataset.}
\centering
\begin{adjustbox}{max width=\textwidth}
\begin{tabular}{c|c|c|cc}
    \hline
    \multirow{2}{4em}{ Datasets } & \multirow{2}{4em}{Task} & \multirow{2}{6em}{NN parameters} & \multicolumn{2}{c}{RS parameters}\\
    \cline{4-5}
        &&&   R & K \\
    \hline
    Adult & classification&  512/256/128  & 500 & 1 \\
     \hline
    phishing & classification &  512/256/128  & 300 & 3 \\
     \hline
    skin & classification &  256/128/64  & 300  & 3  \\
     \hline
    SUSY & classification &  1024/512/256/128/64  & 1000 & 2 \\
     \hline
    abalone & regression &  256/128 & 300 & 1 \\
     \hline
    YearMSD & regression &  1024/512/256/128  & 500 & 3 \\
   
    \hline
	\end{tabular}
	\end{adjustbox}
	\label{table:dataset}
	\vspace{-6mm}
\end{table*}
\subsection{Experimental Setup}
During parameter learning as described by Section~\ref{sec:wholerecipe}, we found setting $M \ll N$ gives good approximations and reduces the kernel computation in training from $O(N^2)$ to $O(N*M)$. Also, we found that learning $x_j$ from a lower dimension suffices. It reduces the training cost substantially. To allow this, we use asymmetric LSH for its flexibility. Simple linear projection is used to keep the hashing cost low during inference. Specifically, for a chosen LSH family $H(x)$ the LSH function for learned vectors is $h = H(x_j)$ while the LSH function for testing data is $h' = H(\mathbf{A}^{\top}x)$. $\mathbf{A}$ is learned along with $x_j$ and $\alpha$. Note these modifications are training time choices. Training on all datasets finished within 1 hour on single Nvidia V100 GPU. We include the cost of the projection matrix in both memory and computation comparison.

\emph{Memory:} We measure the memory based upon number of parameters. All integers and floating point numbers are stored in standard 64-bit. For neural network, the number of parameters is measured by a standard package torchinfo~\cite{torchinfo}. For \textsc{Representer Sketch}, the number of parameters is sum of array ($R*L$) and projection matrix $d*p$.

\emph{Computation:} We measure FLOPs(floating point operations) to compare the computation complexity between \textsc{Representer Sketch} and neural network. For neural network, the number of FLOPs is measured by a standard package~fvcore\cite{fvcore}. For \textsc{Representer Sketch}, we calculate as the sum of projection matrix multiplication, hash computation and aggregation: $2*d*p$ + $\frac{p * K * R}{3}$ + $R$.
\begin{figure}
\vspace{-4mm}
    \begin{minipage}[c]{0.8\textwidth}
        \begin{minipage}[t]{0.4\textwidth}
            \centering
            \subfloat[Adult(higher is better)]{\includegraphics[height=\textwidth]{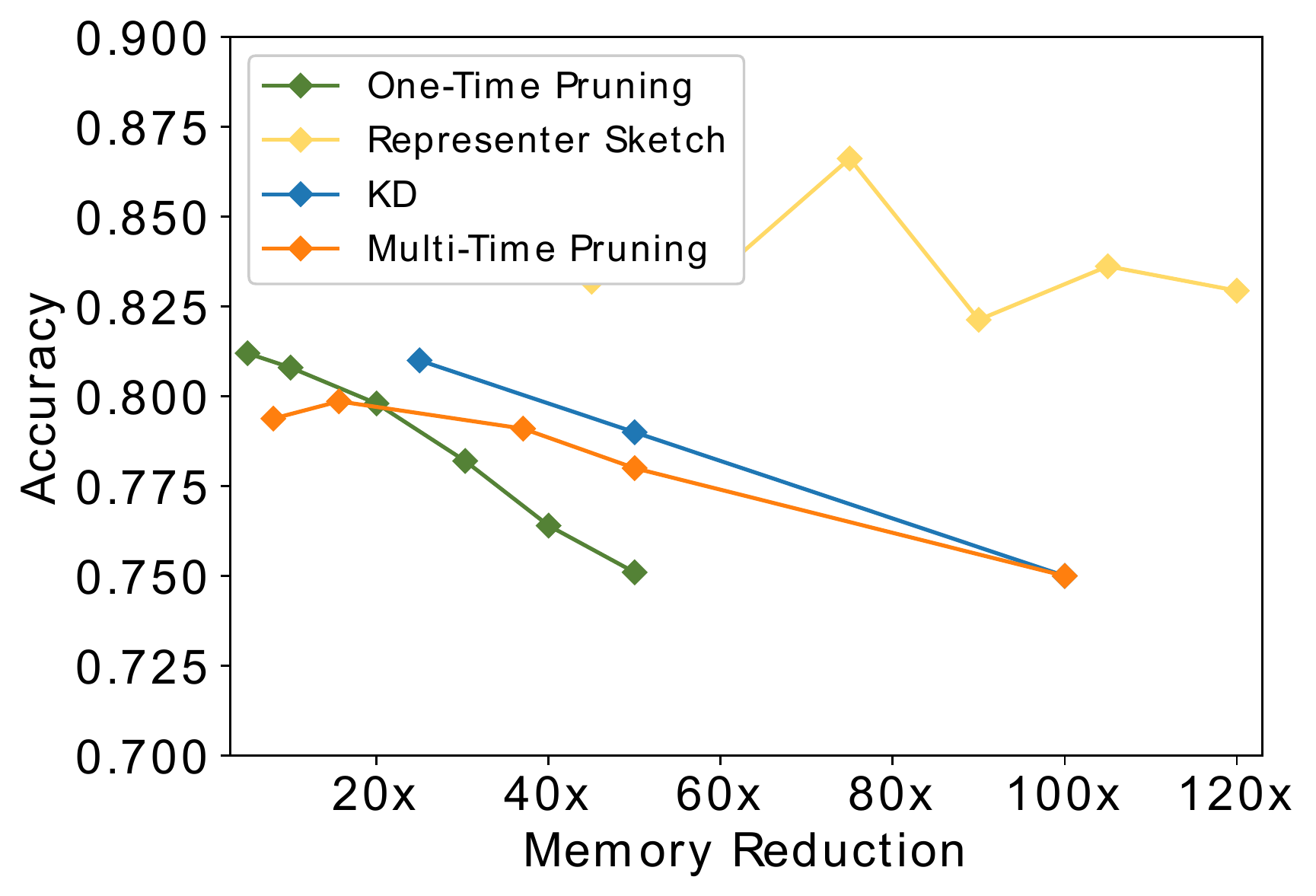}}
        \end{minipage}
        \hfill
        \begin{minipage}[t]{0.4\textwidth}
            \centering
            \subfloat[phishing(higher is better)]{\includegraphics[height=\textwidth]{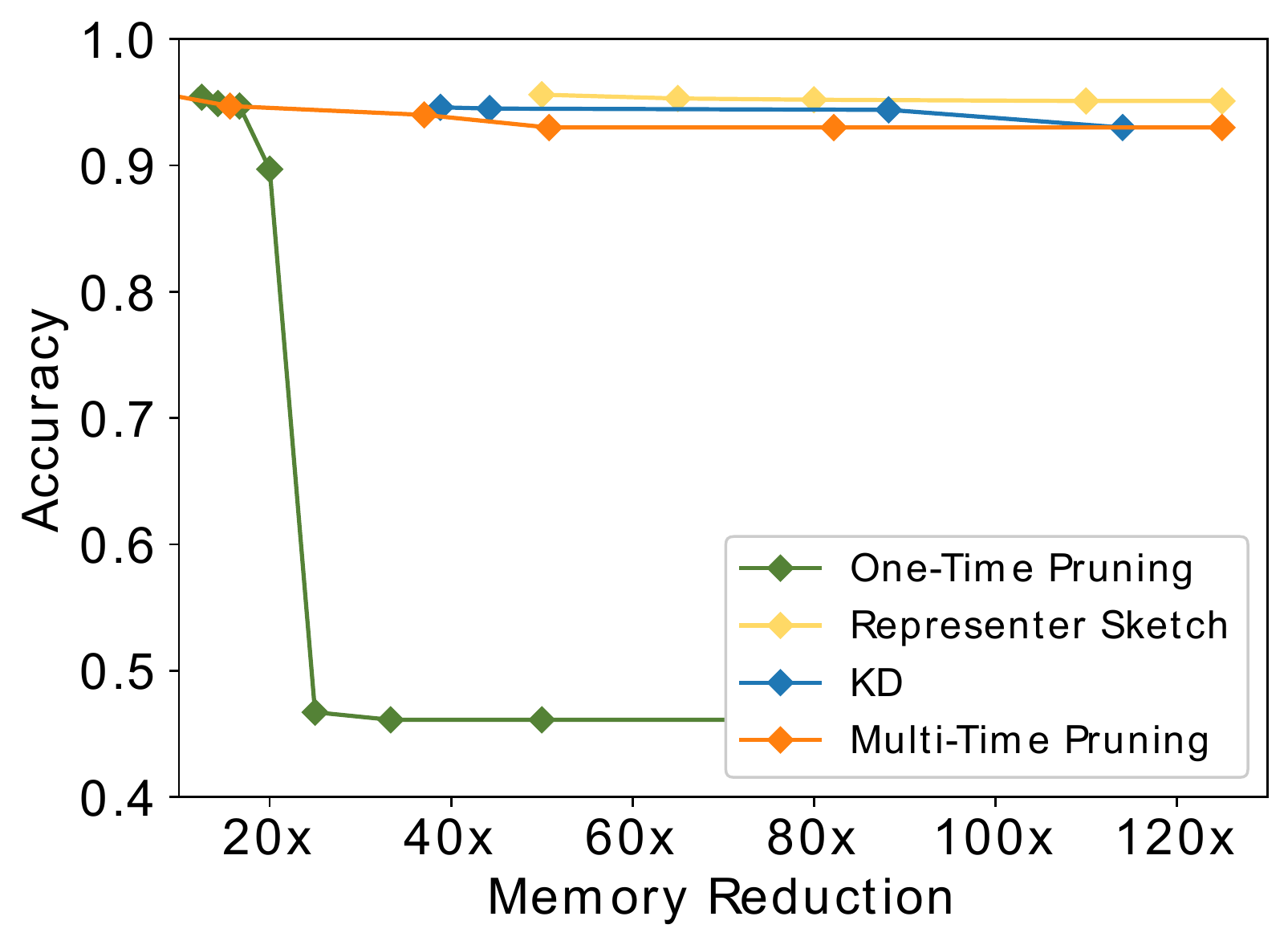}}
        \end{minipage}
    \end{minipage}
    \begin{minipage}[c]{0.8\textwidth}
        \begin{minipage}[t]{0.4\textwidth}
            \centering
            \subfloat[skin (higher is better)]{\includegraphics[height=\textwidth]{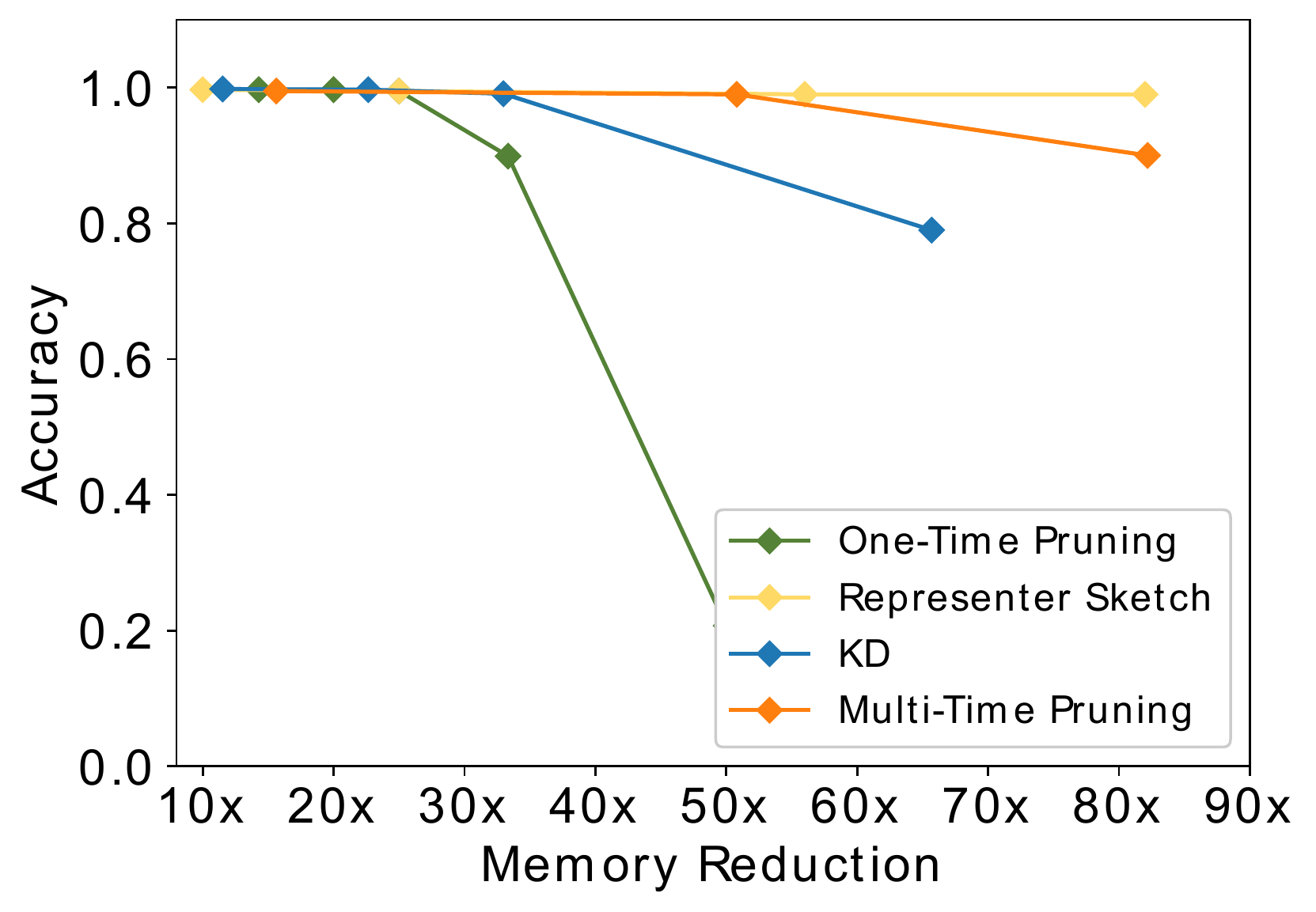}}
        \end{minipage}
        \hfill
        \begin{minipage}[t]{0.4\textwidth}
            \centering
            \subfloat[abalone (Lower is better)]{\includegraphics[height=\textwidth]{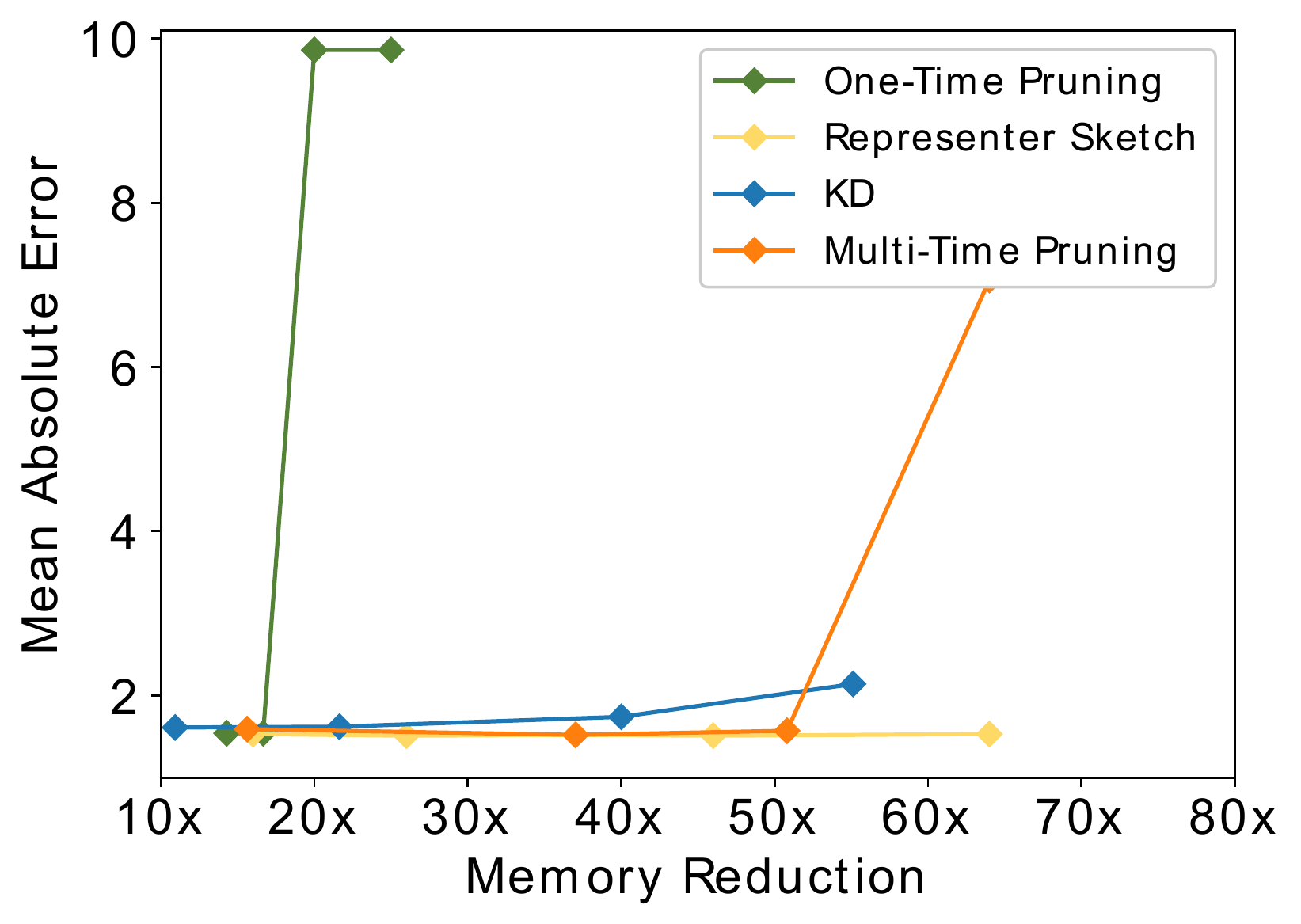}}
        \end{minipage}
    \end{minipage}
\caption{This plot shows the accuracy versus memory reduction rate.  One-Time Pruning refers to pruning and fine-tuning; Multi-Time Pruning refers to pruning and fine-tuning multiple times. KD refers to knowledge distillation. For classification dataset(a)(b)(c), y-axis measures the classification accuracy; higher is better. For regression dataset(d), y-axis measures the mean absolute error; lower is better. X-axis measures how many times each method reduces the memory of a given neural network. \textsc{Representer Sketch} outperforms baselines on all four datasets. We observe that at every level of memory budget, Representer Sketch achieves higher accuracy comparing to baselines and similar accuracy comparing to NN from Table~\ref{table:mainresult}. {\bf The advantage of Representer Sketch far exceeds the memory only comparisons because the inference operations are simpler.}}
\label{fig:accvsmem}
\vspace{-4mm}
\end{figure}
\subsection{Main Results: Saving Inference Cost}
Table\ref{table:mainresult} summarizes the accuracy, memory consumption, and computation complexity of \textsc{Representer Sketch}. In general, \textsc{Representer Sketch} achieves at least competing accuracy while greatly reduce memory requirement and computation complexity. NN refers to a trained neural network, Kernel refers to its weighted kernel density function, and RS refers to \textsc{Representer Sketch}. For the classification task, accuracy is measured as standard classification accuracy. For the regression task, accuracy is measured as mean absolute error. Parameter settings can be found in Table~\ref{table:dataset}. 

We observe that Kernel consistently achieves competing or higher accuracy comparing to the trained neural network. This confirms our theoritical result in Section~\ref{sec:33} that a weighted LSH kernel density representation can achieve satisfying performance. Besides, we observe that \textsc{Representer Sketch} obtains similar accuracy comparing against Kernel. This shows that our sketching algorithm can accurately approximate a weighted kernel density function. We observe some approximation noises on datasets such as SUSY and YearMSD. The approximation error can influence the performance in good and bad ways: On YearMSD, the accuracy increases; On SUSY, the accuracy drops. 

From an efficiency perspective, \textsc{Representer Sketch} achieves 17.8x to 114x memory reduction and 4x to 59x reduction in FLOPs. We observe that the memory of \textsc{Representer Sketch} is on the scale of kilobytes, which is small enough to fit in local memory and edge devices. Also, we observe a dramatic drop in FLOPs for \textsc{Representer Sketch}. This confirms the low computation complexity for our algorithm as it consists of mainly addition and subtraction. 


\subsection{Memory Foot-Print Comparison with Iterative Pruning and Knowledge Distillation}
In this section, we compare \textsc{Representer Sketch} with pruning and knowledge distillation on the accuracy versus memory trade-off. Details can be found in Section~\ref{sec:baseline}. It should be noted that we anticipate \textsc{Representer Sketch} advantage to be even more prominent than just the memory benefit because the computations are much simpler than other baselines. \textsc{Representer Sketch} mostly relies on simple addition subtraction operation, whereas the baselines are still reduced versions of neural networks with a lot of multiplications which are about 4x more costly. 

From Figure \ref{fig:accvsmem}, we observe that \textsc{Representer Sketch} outperforms baselines on all datasets. At a low level of memory reduction rate, pruning and knowledge distillation achieves comparable accuracy. We observe a steep drop in classification accuracy and a sudden increase in regression error for baselines as we reduce the memory budget. However, with \textsc{Representer Sketch}, we observe the line remains flat. For a tighter memory budget, \textsc{Representer Sketch} is a better choice as it maintains high accuracy even with over 100 times memory reduction.

\subsection{Discussion on Limitation and Negative Social Impact }
\label{sec:limitation}
Our algorithm mainly concerns that efficiency problem with current neural network inference, thus we believe that it does not pose potential negative social impact. There exists one limitation of \textsc{Representer Sketch} in multi-class classification task: the sketch size, even though small, asymptotically grows linearly with the number of classes. However, we believe this issue can be mitigated and we leave it for further work.

%% file: conclusion.tex
\vspace{-2mm}
\section{Conclusion}

In this work, we present \textsc{Representer Sketch} to extract knowledge from a neural network to a concise set of weighted count arrays for the purpose of efficient inference. With theoretical support, we transform a neural network function to its weighted LSH kernel density representation, which then can be accurately approximated with a small set of array lookups. \textsc{Representer Sketch} reduces the memory requirement and computation complexity of inference without affecting accuracy. The smaller storage requirement and simpler computation operations make inference more energy-efficient and more friendly for deployment on resource-constrained devices such as sensors and embedded systems.

%% file: appendix.tex
\section{Appendix}

\subsection{Theory}
In this section, we provide complete proofs for our theoretical results.
\input{proof}

%% file: proof.tex
\subsubsection{Proof of Theorem \ref{thm:expandvar}}
\begin{customthm}{1}
Given a dataset $\mathcal{D}$ of weighted samples $\{(\alpha_{x_i},x_i)\}$, let $h(x)$ be an LSH function drawn from an LSH family with collision probability $\mathcal{K}(\cdot,\cdot)$. Let S be a row of the sketch constructed using $h(x)$. For any query $q$,
\begin{equation*}
    \mathbb{E}(S[h(q)]) = \sum_{i}^{|\mathcal{D}|} \alpha_{x_i} \mathcal{K}(x_i,q), \qquad
    \mathrm{var}\left(S[h(q)]\right) \leq \left( \sum_{i}^{|\mathcal{D}|} \alpha_{x_i} \sqrt{\mathcal{K}(x_i,q)} \right)^2
\end{equation*}
\end{customthm}
\begin{proof} Let $\mathds{1}_i$ denote the indicator function  $\mathds{1}_{h_(x_i) = h_(q)}$. That is, $ \mathds{1}_i = 1$ when data $x_i$ from the dataset collides with the query $q$.  To simplify the presentation, let $Z = S[h(q)]$. 

\textbf{Expectation:} The value in the sketch can be written as
$$Z = \sum_{i}^{|\mathcal{D}|} \alpha_{x_i}\mathds{1}_i$$
By the linearity of the expectation
$$\mathbb{E}(Z) = \sum_{i}^{|\mathcal{D}|} \alpha_{x_i}\mathbb{E}(\mathds{1}_i)$$
We know that $\mathbb{E}(\mathds{1}_i)$ is the collision probability of $h(x)$, thus,
$$\mathbb{E}(Z) =\sum_{i}^{|\mathcal{D}|} \alpha_{x_i}\mathcal{K}(x_i,q)$$

\textbf{Variance:} The variance is bounded by the second moment. The second moment of this estimator can be written as
$$\mathbb{E}(Z^2) =\sum_{i}^{|\mathcal{D}|} \sum_{j}^{|\mathcal{D}|}\alpha_{x_i}\alpha_{x_j}\mathbb{E}(\mathds{1}_i\mathds{1}_j)$$
By the Cauchy-Schwarz inequality,
$$\mathbb{E}(\mathds{1}_i\mathds{1}_j) \leq \sqrt{\mathbb{E}(\mathds{1}_i)\vphantom{\mathbb{E}(\mathds{1}_j)}} \sqrt{\mathbb{E}(\mathds{1}_j)}$$
Then,
$$\mathbb{E}(Z^2) \leq \sum_{i}^{|\mathcal{D}|} \sum_{j}^{|\mathcal{D}|}\alpha_{x_i}\alpha_{x_j}\sqrt{\mathcal{K}(x_i,q)\vphantom{\mathcal{K}(x_j,q)}}\sqrt{\mathcal{K}(x_j,q)} = \left( \sum_{i}^{|\mathcal{D}|} \alpha_{x_i} \sqrt{\mathcal{K}(x_i,q)} \right)^2$$
Thus,
$$\mathrm{var}(Z) \leq \left( \sum_{i}^{|\mathcal{D}|} \alpha_{x_i} \sqrt{\mathcal{K}(x_i,q)} \right)^2$$
\end{proof}

\subsubsection{Proof of Lemma \ref{lem:med}}
\begin{customlemma}{1}
Let $Z_1,...Z_R$ be $L$ i.i.d. random variables with mean $\mathbb{E}[Z] = \mu$ and variance $\leq \sigma^2$. Divide the $L$ variables into $g$ groups so that each group contains $m = L / g$ elements, and take the empirical average within each group. The median-of-means estimate $\hat{\mu}$ is the median of the $g$ group means. If $g = 8 \log(1/ \delta)$ and $m = L / g$, then the following statement holds with probability $1 - \delta$.
$$ |\hat{\mu} - \mu|\leq 6\frac{\sigma}{\sqrt{L}} \sqrt{\log{1/\delta}} $$
\end{customlemma}
\begin{proof}
This proof is given in \cite{alon1999space} as the proof of Theorem 2.1 (which is a slightly more general version of the statement above).
\end{proof}


\subsubsection{Proof of Theorem \ref{thm:mem_error}}
\begin{customthm}{2}
Let $Z(q)$ be the median-of-means estimate constructed using the $L$ unbiased estimators of the sketch with $L$ rows. Then with probability $1 - \delta$,

\begin{equation*}
    |Z(q) - f_K(q)| \leq 6\frac{ \tilde{f}_K(q)}{\sqrt{L}}\sqrt{\log{1/\delta}}
\end{equation*}
where $f_K(q)$ and $\tilde{f}_K(q)$ are the weighted KDE with kernels $\mathcal{K}(x,q)$ and $\sqrt{\mathcal{K}(x,q)}$, respectively.
\end{customthm}

\begin{proof}
From Theorem \ref{thm:expandvar}, we know that 
 $$\sigma \leq \sum_{i}^{|\mathcal{D}|} \alpha_{x_i} \sqrt{\mathcal{K}(x_i,q)}$$
 
 Substituting this variance bound into Lemma \ref{lem:med} proves the theorem.
\end{proof}

\subsubsection{Proof of Lemma \ref{lemma2}}
\begin{customlemma}{2}
The L2 LSH kernel from~\cite{coleman2019sublinear} is shift-invariant and universal.
\end{customlemma}
\begin{proof}
To see that the kernel is shift-invariant, observe that $K(x,y) = K(\mathrm{dist}(x,y))$ and that the Euclidean distance $\|x - y\|_2$ is a function of the difference $x - y$. Thus, $K(x,y) = K(x - y)$.

Since the kernel is shift-invariant, it is sufficient to show that the support of the Fourier transform of the kernel is the entire real line~\cite{carmeli2010vector}. Observe from~\cite{datar2004locality} that the kernel may be written as
$$ k(c) = \int_{0}^r \left(1 - \frac{t}{r}\right)\frac{1}{c} e^{-\frac{t^2}{2c^2}}dt$$
where $c = \mathrm{dist}(x,y)$ and $r$ is a parameter. The Fourier transform of this quantity is 
$$ \int_{-\infty}^{\infty}\left( \int_{0}^r \frac{2}{\sqrt{2\pi}}\left(1 - \frac{t}{r}\right)\frac{1}{c} e^{-\frac{t^2}{2c^2}} dt\right) e^{i\omega c} d\omega $$
The integrand satisfies the requirements of Fubini's Theorem (absolutely integrable), so we may exchange the order of integration.
$$ \int_{0}^r \frac{2}{\sqrt{2\pi}}\left(1 - \frac{t}{r}\right) \left(\int_{-\infty}^{\infty} \frac{1}{c} e^{-\frac{t^2}{2c^2}}  e^{i\omega c} d\omega \right)dt$$
Observe that the Fourier transform of the inner integrand has full support because the exponential function is nonzero everywhere. Now observe that the outer integral is the limit of a sum of Fourier transforms, each of which has the real line as its support.
Because $\frac{2}{\sqrt{2\pi}}(1 - t/r) > 0$ over the integration region, the coefficients of this sum are positive. Therefore, the Fourier transform of the kernel has the real line as its support.
\end{proof}

\subsubsection{Proof of Theorem \ref{thm:lshrepresenter}}
\begin{customthm}{3} Given a continuous and bounded function $g(q)$ and $\epsilon > 0$, there exists a set of coefficients $\{\alpha_n\}$, set of points $\{x_n\}$ and an integer $N$ such that 
$$f_N(q) = \sum_{n=1}^N \alpha_N \mathcal{K}(x_n,q) \qquad \|f_N(q) - g(q)\|_\mathcal{X} \leq \epsilon$$
where $\mathcal{K}(x_n,q)$ is the L2 LSH kernel and $\mathcal{X}$ is any compact subset of $\mathbb{R}^{d}$.
\end{customthm}

\begin{proof}
This follows directly from the definition of a universal kernel.
\end{proof}

\subsubsection{Proof of Corollary \ref{cor:transform}}
\begin{customcorollary}{1}
Given an injective transform $T(q): \mathbb{R}^{d} \mapsto \mathbb{R}^{d'}$ and a continuous and bounded function $g(q): \mathbb{R}^{d} \mapsto \mathbb{R}$, there exists a set of points $\{x_n\}$ in the \textit{transformed space} $\mathbb{R}^{d'}$, coefficients $\{\alpha_n\}$, and an integer $N$ such that 
$$f_N(q) = \sum_{n=1}^N \alpha_n \mathcal{K}(x_n,T(q)) \qquad \|f_N(q) - g(q)\|_S \leq \epsilon$$
\end{customcorollary}

\begin{proof}Construct a function $p(x)$ such that 
$$
 p(x) = \begin{cases}
         g(q) & \text{for } x = T(q)\\
         0, & \text{otherwise } \end{cases} $$
Because $T(q)$ is injective, every value of $q$ maps to a unique value of $q' \in \mathbb{R}^{d'}$. Note that we may query $p(x)$ with $x = T(q)$ to get the value of $g(q)$ (i.e. $g(q) = p(T(q))$). Therefore, any approximation to $p(x)$ also yields an approximation to $g(q)$ when we restrict the domain to points $x: x = T(q)$. Also, note that if we construct an approximation to $p(x)$ having $\epsilon$ integrated error, we obtain an approximation to $g(q)$ having $\leq \epsilon$ integrated error because the error of the second approximation is computed over the (smaller) subspace $x : x = T(q)$. 

By Theorem \ref{thm:lshrepresenter}, there is a set of coefficients $\{\alpha_n\}$, set of points $\{x_n\}$ and an integer $N$ such that 
$$f_N(q) = \sum_{n=1}^N \alpha_N \mathcal{K}(x_n,q) \qquad \|f_N(q) - p(T(q))\|_\mathcal{X} \leq \epsilon$$

As noted before, this provides a $\leq \epsilon$ error approximation to $g(q)$, proving the corollary. 

\end{proof}